%% file: powerRemainder.tex
\documentclass[envcountsame]{llncs}

\input{style}

\title{Scattered Factor Universality - The Power of the Remainder}
\titlerunning{The Power of the Remainder}
\author{Pamela Fleischmann\inst{1}\thanks{Supported by DFG grant 437493335}\and 
Sebastian Bernhard 
Germann\inst{1}
\and Dirk Nowotka\inst{1}}
\authorrunning{P. Fleischmann \and S.B. Germann \and D. Nowotka}

\institute{Kiel University, 
Germany\\ \email{fpa@informatik.uni-kiel.de,
stu121684@mail.uni-kiel.de,dn@informatik.uni-kiel.de}}

\newif\ifpaper
\paperfalse 

\begin{document}

\maketitle
\begin{abstract}
\input{abstract}
\end{abstract}

\section{Introduction}
\input{intro}

\section{Preliminaries}\label{prel}
\input{prelims}

\section{The Growth Behaviour of the Universality}\label{univ:sec:powrem}
\input{growth}

\section{Chaining the Remainder} \label{univ:sec:chaintherem}
\input{chain}

\section{Conclusion}\label{conlcusion}
\input{conclusion}

\bibliographystyle{plain}
\bibliography{powerRemainder}

\appendix

\clearpage
\ifpaper 
	\section{Appendix: Proofs}
	\label{sec:proof}
    \input{appendix-proofs}
 \else    
 \fi

\end{document}

%% file: style.tex
\usepackage[utf8]{inputenc}
\usepackage[T1]{fontenc}
\usepackage[ngerman, british]{babel}
\usepackage{amsmath}
\usepackage{amsfonts}
\usepackage{amssymb}
\usepackage{authblk}
\usepackage{mathpazo} 
\usepackage{hyperref}
\usepackage{todonotes}
\usepackage{braket} 

\newcommand{\N}{\mathbb{N}}
\newcommand{\Nz}{\mathbb{N}_0}

\newcommand{\alphabet}{\Sigma}
\newcommand{\asize}{\sigma}
\newcommand{\alphs}{\Sigma^*}
\newcommand{\alphp}{\Sigma^+}

\newcommand{\ta}{\mathtt{a}}
\newcommand{\tb}{\mathtt{b}}
\newcommand{\tc}{\mathtt{c}}

\newcommand{\tn}{\mathtt{n}}
\newcommand{\emptyword}{\varepsilon}
\newcommand{\univ}{\iota}
\newcommand{\circuniv}{\zeta}
\newcommand{\rem}{r}
\newcommand{\modus}{m}

\newcommand{\bigo}{\mathcal{O}}
\newcommand{\simonsc}{\sim_k}

\newcommand{\alfrem}[1]{\mathcal{R}( #1 ) }

\newcommand{\bdiff}[1]{\nabla #1}
\newcommand{\phaseout}[1]{}
\newcommand{\abs}[1]{\left\lvert #1 \right\rvert}
\newcommand{\disinv}[2]{\left[ #1 , #2 \right]}

\DeclareMathOperator{\alf}{alph}

\DeclareMathOperator{\arch}{ar}
\DeclareMathOperator{\scatfact}{ScatFact}

%% file: abstract.tex
Scattered factor (circular) universality was firstly introduced by Barker et al. in 2020. A word 
$w$ is called $k$-universal for some natural number $k$, if every word of length $k$ of $w$'s 
alphabet occurs as a scattered factor in $w$; it is called circular $k$-universal if a conjugate of 
$w$ is $k$-universal. Here, a word $u=u_1\cdots u_n$ is called a scattered factor of $w$ if $u$ is 
obtained from $w$ by deleting parts of $w$, i.e. there exists (possibly empty) words 
$v_1,\dots,v_{n+1}$ with $w=v_1u_1v_2\cdots v_nu_nv_{n+1}$. In this work, we prove two problems, 
left open in the aforementioned paper, namely a generalisation of one of their main theorems to 
arbitrary alphabets and a slight modification of another theorem such that we characterise the 
circular universality by the universality. On the way, we present deep insights into the behaviour 
of the remainder of the so called arch factorisation by Hebrard when repetitions of words are 
considered.

%% file: intro.tex
By deleting letters from a word one obtains another word, a so called {\em scattered 
factor} (also known as \emph{subword} or \emph{subsequence}). More 
formally, a word $u=u_1\cdots u_n$ is a scattered factor of $w$ if there exist 
(possibly) empty words $v_1,\dots,v_{n+1}$ with $w=v_1u_1\cdots v_nu_nv_{n+1}$.
For instance, $\mathtt{latin}$ is a scattered factor of $\mathtt{dalmatian}$ but $\mathtt{lama}$ is 
not.
Scattered factors are a fundamental concept in mathematics and computer 
science: whenever data are transmitted via a lossy channel or in aligning 
DNA-sequences, scattered factors are the formal model to describe the incomplete data
(e.g. \cite{ElzingaRW08}). 
Thus, it is not astonishing that scattered factors are strongly related to partial 
words  \cite{blanchetpartial2012}. Moreover, \emph{automatic sequences} 
can be in part understood by their \emph{subword complexity}, a measure of complexity 
on words that is defined by scattered factors 
\cite{allouche_shallit_2003}. \emph{Parikh matrices} and {\em subword histories} use 
scattered factors to encode numerical properties of words into matrices, thus 
connecting the world of words and languages with the world of vectors and matrices 
(see \cite{Mat04,Salomaa05,Seki12}). From an algorithmic point of view, scattered 
factors are crucial in some classical problems: the longest common subsequence, the 
shortest common supersequence, and the string-to-string correction problem 
\cite{Maier:1978,BringmannK18,Wagner:1974}. On the other hand, 
scattered factors are
also used in logic-theories and have applications in formal software verifcation
\cite{Zetzsche16,HalfonSZ17,KuskeZ19}.

The line of research that lead to this work began by Higman 
\cite{Higman1952}, who showed that in any infinite set of words 
there are always two words such that one is a scattered factor of the other, 
albeit only as an application of a more general theorem about partial 
orderings on an abstract algebra and without explicitly defining scattered 
factors. Later in 1967 Haines \cite{Haines1969OnFM} explicitly introduced scattered 
factors and rediscovered Higman's result. In the seminal work 
\cite{Simon1975PiecewiseTE} from 1975, Simon  used this partial ordering to define 
the equivalence relation $\simonsc$, now known as {\em Simon congruence}, where $x 
\simonsc y$ iff $x$ and $y$ have the same set of scattered factors of a fixed 
length $k$. In 1991, Hebrard introduced the {\em arch factorisation} which is a very powerful tool 
in investigating the scattered factors of a word \cite{HEBRARD199135}. A very profound overview 
from a mathematical point of view can be found in \cite[Chapter 6]{lothaire_one}, where Simon and 
Sakarovitch expand Simon's previous work.

In this work we focus on a special $\simonsc$-class of words. A word $w 
\in \alphs$ is called \emph{$k$-universal} if its set of scattered factors of 
length $k$ is $\alphabet^k$. For instance, $\mathtt{anana}$ is $2$-universal over 
$\{\ta,\tn\}$ but $\mathtt{banana}$ is only $1$-universal over $\{\ta,\tb,\tn\}$.
Notice, that this notion is equivalent to the notion of {\em richness} introduced in 
\cite{CSLKarandikarS,journals/lmcs/KarandikarS19} in the context of piecewise 
testable languages; as in \cite{barker2020scattered} we prefer the notion of universality for 
avoiding confusion with the richness w.r.t. palindromes. While the classical universality problem, 
which asks whether a given language $L\subseteq\Sigma^{\ast}$ is equal to $\Sigma^{\ast}$, and many 
variants of it, as well as the universality problem for (partial) words, which asks given an 
$\ell$ whether there exists a word $w\in\Sigma^{\ast}$ that contains all words of length $\ell$ 
exactly once as a factor are well studied
(see \cite{HolzerK11,GawrychowskiRSS17,Rampersad:2012,KrotzschMT17} and 
\cite{martin1934,Bruijn46,ChenKMS17,GoecknerGHKKKS18} and the references therein) , the 
universality problem for scattered factors just got recently attention 
(see 
\cite{fleischmannweaklycbalanced2019,barker2020scattered,fleischmann2020edit,GawrychowskiKKM21} and 
the references therein).

\textbf{Our contribution.} Following the line of research started in \cite{barker2020scattered},
we investigate  the (circular) universality of words. In particular, we study the 
universality of repetitions, which leads to several characterisations of its 
growth by the remainder of the arch factorisation. We present that intervals, on which the 
universality of repetitions is constant, correspond to either ascending or 
descending chains of the remainder of those repetitions. These deep insights into the behaviour of 
the remainder of the arch factorisation are linked to the circular universality such that we are 
able to present results on two open problems of \cite{barker2020scattered}. As a 
consequence, we also get an efficient algorithm to compute the circular 
universality of a word.

\textbf{Structure of the work.}
In Section~\ref{prel} we present the basic notions and in Section~\ref{univ:sec:powrem}
we study the remainder of the arch factorisation and especially its growth behaviour on repetitions.
Afterwards, in Section~\ref{univ:sec:chaintherem}, we connect the previous results and define 
ascending and descending chains of the remainders which leads to our main results, the 
generalisations of Theorem~22  and Theorem~23 from
\cite{barker2020scattered}.

\ifpaper
Due to space restrictions, some proofs (marked with $\ast$) can be found in the appendix.
\else
\fi

%% file: prelims.tex
Let  $\N = \set{1,2,\ldots }$ denote the natural numbers. Set $\N_0 = \N 
\cup \set{0}$ and $\N_{\geq k} = \set{n \in \N | n \geq k}$  for a $k\in\N_0$.
We 
also define the 
discrete interval $\disinv{i}{j} = \set{i, i+1, \ldots, j}$ for 
$i,j\in\N_0$. Define for a function $f: \Nz \rightarrow \Nz$ 
the \emph{backward difference} in $x\in\N$ by $\bdiff{f(x)} = f(x) - f(x-1)$ and 
call $\bdiff$ the \emph{backward difference operator} and $\bdiff{f(x)}$ the 
\emph{growth} of $f$ in $x$.

An \emph{alphabet} $\alphabet$ is a finite set of symbols, called 
\emph{letter}.  A \emph{word} $w$ is a finite 
sequence of letters from a 
given alphabet and its length $|w|$ is the number of $w$'s letters. For $i \in 
\disinv{1}{\abs{w}}$ let $w[i]$ 
denote the $i^{\mathrm{th}}$ letter of $w$. The set of all 
finite words over the alphabet $\alphabet$, 
denoted by $\alphs$, is the free monoid generated by $\alphabet$ with 
concatenation (product) as operation and the neutral element is the empty word 
$\varepsilon$, i.e. the word of length $0$. Set $\alphp = \alphs \setminus \set{ 
\emptyword }$ and $\alphabet^k = \set{ w \in \alphs | 
\abs{w} = k}$ for some $k \in \N$. Let $u,w\in\alphs$ be words. Then $u$ is called a \emph{factor} 
of $w$, if $w 
= xuy$ for some words $x$ and $y$ over $\alphabet$. If $x = \emptyword$ (resp. 
$y = \emptyword$) then $u$ is called a \emph{prefix} (resp. \emph{suffix}) of 
$w$. A factor (resp. prefix, suffix) $u$ of $w$ is called a \emph{proper} 
factor (resp. prefix, suffix), if $u \notin \set{w, \emptyword}$. If $w = xy$ 
then we define $x^{-1}w = y$ and $wy^{-1} = x$. Furthermore $u$ and $w$ are 
called \emph{prefix-compatible} (resp. \emph{suffix-compatible}) if one is a prefix 
(resp. suffix) of the other. Two words $w$ and $u$ are said to be \emph{conjugate} to 
each other if there 
exist words $x,y \in \alphs$ such that  $w = xy$ and $u = yx$.
We denote the \emph{reversal} of a word by $w^R$, i.e. if $\abs{w} = n$ then 
$w^R = w[n] \cdot w[n-1] \cdot \ldots \cdot w[1]$. We say that a letter $\ta 
\in \alphabet$ \emph{occurs} in $w$, if $\ta$ is a factor of $w$. We denote 
the set of all letters that occur in $w$ by $\alf(w)$.
If we have words $w_i\in\alphs$ for all $i\in \disinv{1}{n}$ and some $n\in\N$, 
then we define $\prod_{i=k}^n w_i = w_k \cdot \ldots \cdot w_n$. In the 
special case $w_1 = w_2 = \ldots = w_n =: w$ we also write $w^n= 
\prod_{i=1}^{n} w$ and call $w^n$ the $n^{\mathrm{th}}$ power of $w$.

Now, we introduce the basic notions around scattered factor universality, firstly introduced in 
\cite{barker2020scattered}.

\begin{definition}
    A word $u \in \alphs$ is called a \emph{scattered factor} (or 
    \emph{subword}) of a word $w\in \alphs$ if there exist 
    $x_1, \ldots, x_{|u|+1} \in \alphs$ such that
    $w = x_1 u[1] x_2 \cdots x_{|u|} u[|u|] 
    x_{|u|+1}$.
    We denote the set of all scattered factors of a given word $w\in\Sigma^{\ast}$ 
    by 
    $\scatfact(w)$ and the set of all scattered factors of a given length 
    $k\in\N_0$ by $\scatfact_k(w)$.
\end{definition}

\begin{definition}
    A word $w \in \alphs$ is called $k$-\emph{universal} (w.r.t $\alphabet$) 
    for some $k\in\N_0$ if $\scatfact_k(w) = \alphabet^k$. 
    A word $w \in \alphs$ is called \emph{circular} $k$-universal (w.r.t. 
    $\alphabet$) for some $k \in \N_0$, if there exists a conjugate $v$ of $w$ 
    such that $v$ is $k$-universal. 
    We define the 
    \emph{universality index} of $w$ as the maximal $k$ such that $w$ is 
    $k$-universal and denote it by $\univ(w)$; analogously defined, $\circuniv(w)$ 
    denotes the {\em circular universality index} of $w$.
\end{definition}

\begin{remark}
By definition, the universality is w.r.t. to a given alphabet $\Sigma$. Notice, that 
we have immediately $\iota(w)=0=\zeta(w)$ if $\alf(w)\subsetneq\Sigma$. Therefore, 
we implicitly assume w.l.o.g. $\Sigma=\alf(w)$ from now on. Thus, we also assume 
$\iota(w)>0$ at any time without mentioning it. For abbreviation, we set 
$\sigma=|\Sigma|$.
\end{remark}

Since this work focuses on the (circular) universality index of powers of 
$w\in\Sigma^{\ast}$, we introduce the following 
parametrisation.

\begin{definition}
For $w\in\alphs$ define $\univ_w : \N_0 \rightarrow \N_0;\,s \mapsto \univ(w^s)$
and $\circuniv_w : \N_0 \rightarrow \N_0;\,s \mapsto \circuniv(w^s)$.
\end{definition}

The following remark captures properties of $\iota_w$ and $\zeta_w$ for a given 
$w\in\Sigma^{\ast}$.

\begin{remark}    \label{univ:rem:boundtheuniversality}
Consider $w \in \alphs$ with $k = \univ(w)$. Then we have, for all $s\in\N_0$,
first, $\univ(w^s) = \univ_w(s) = \sum_{i=1}^{s} \bdiff{\univ_w(i)}$, second
$k \leq \bdiff{\univ_w(s)} \leq k+1$, and finally $\bdiff{\univ_w(1)} = k$. Therefore, 
$sk \leq \univ_w(s) \leq sk + s - 1$. There are equivalent ways to represent 
these two extreme cases, 
        namely we have\\
        1. $\univ_w(s) = sk$ iff $\bdiff{\univ_w(i)} = k$ for all $i 
            \in \disinv{1}{s}$ and \\
        2. $\univ_w(s) = sk + s - 1$ iff $\bdiff{\univ_w(i)} = k+1$ for 
            all $i \in \disinv{2}{s}$.
\end{remark}

Next we recall the arch factorisation introduced by Hebrard in 
\cite{HEBRARD199135}, which is a powerful tool in investigating the universality.

\begin{definition}
    Let $w \in \alphs$. The factorisation $w = \arch_1(w) \cdot \ldots \cdot 
    \arch_k(w) \cdot \rem(w)$, for some $k\in\N_0$, is called \emph{arch 
factorisation} if for all 
    $i \in \disinv{1}{k}$ we have $\arch_i(w) = u_i \ta_i$ for some $u_i \in \alphs$ 
and $\ta_i \in \alphabet$, $\alf(u_i) \neq \alphabet$, $\alf(\arch_i(w)) = 
\alphabet$,
and $\alf(\rem(w)) \neq \alphabet$.
    Furthermore, we define $\modus(w) = \ta_1 \cdot \ldots  \cdot \ta_k$.
    We call $\arch_i(w)$ the $i^{\mathrm{th}}$ arch, $\rem(w)$ the remainder, 
    and  denote the set of letters that occur in the remainder 
    by $\alfrem{w} = \alf(\rem(w))$.
\end{definition}

\begin{remark}
Note, that for all $w\in\Sigma^{\ast}$ and appropriate $i\in\N$, $m(w)[i]$ is unique 
in $\arch_i(w)$ and the number of arches of a word $w$ is $\univ(w)$. In 
\cite[Proposition 10]{barker2020scattered} the arch factorisation
was computed recursively by: $\arch_1(w)$ as the shortest prefix of $w$ 
        such that $\alf(\arch_1(w)) = \alphabet$ and  then $\arch_i(w) = 
        \arch_{1}((\arch_1(w)\cdots \arch_{i-1}(w))^{-1} w)$. 
\end{remark}

In examples we will visualise the arch factorisation with the 
        use of brackets. For example, we will write $
        (\ta\tb\tc) \cdot (\tc\tb\tb\ta) \cdot (\tc\ta\ta\ta\tb) \cdot \ta$
        to indicate the three arches $\ta\tb\tc,\tc\tb\tb\ta$, and $\tc\ta\ta\ta\tb$
        and the remainder (without brackets) $\ta$.

In \cite{barker2020scattered} the dynamic between universality and circular 
universality is studied. Na\"{i}vely, one would expect that the universality of 
powers of $w$ grows linearly with its universality, i.e. $\nabla{\univ(w^s)} = 
\univ(w)$ for all $s\in\N$. But this is not always the case.  Instead, its 
actual growth is related to the circular universality of $w$. 
The following statements about (circular) universality from 
\cite{barker2020scattered} are fundamental and the basis for our work.

\begin{lemma}[\cite{barker2020scattered}]
    \label{lem:univ001}
    For $w \in \alphs$, we have $\univ(w)=\univ(w^R)$ and $\iota(w) \leq \circuniv(w) 
\leq \iota(w) +1$.
\end{lemma}

\begin{theorem}[\cite{barker2020scattered}]
    \label{theo:barker22}
    Let $w\in\alphs$ and $k=\univ(w)$. For all $s\in\N$, if $\circuniv(w)=k+1$ 
    then $\univ(w^s) = sk+s-1$.
\end{theorem}

\begin{theorem}[\cite{barker2020scattered}]
    \label{theo:barker23}
    Let $\abs{\alphabet} = 2$, $w\in\alphs$ with $k=\univ(w)$, and $s \in \N$. 
    Then $\univ(w^s) = sk+s-1$ if $\circuniv(w) = k+1$ and $\univ(w^s) = sk$ 
    otherwise.
\end{theorem}

As stated in \cite{barker2020scattered}, neither Theorem~\ref{theo:barker23} nor
the converse of Theorem~\ref{theo:barker22} hold for ternary alphabets witnessed by 
the following example: considering $ w= (\tb\ta\tb\tc) \cdot (\tc\ta\ta\tb) \cdot \tc$,
we have $\circuniv(w)=\iota(w)$ and $\univ(ww)=2\iota(w)+1$. 

We finish the preliminaries with a lemma that follows the line of arguments 
repeatedly used in \cite{barker2020scattered}.
\ifpaper
\begin{lemma}[$\ast$]
\else
\begin{lemma}
\fi
    \label{univ:lem:circtonorm}
    Let $w \in \alphs$ and $k=\univ(w)$. If $\bdiff{\univ_w(s)} = 
    k$ for all $s \in \N$ then also $\bdiff{\circuniv_w(s)} = k$ for all $s 
    \in \N$.
\end{lemma}

\ifpaper 
\else    
\input{proof_circtonorm}
\fi

%% file: proof_circtonorm.tex
\begin{proof}
    Let $\bdiff{\univ_w(s)} = k$ for all $s \in \N$. Now suppose that 
    $\bdiff{\circuniv_w(s_0)} = k+1$ for some $s_0 \in \N$ and choose $s_0$ to 
    be minimal with this property. Then we have $\circuniv_w(s_0) = s_0 k + 
    1$. This implies that $\circuniv_w(2s_0) \geq 2s_0 k + 2$. Therefore we 
    get $\univ_w(2s_0) \geq 2s_0 + 1$ by Lemma~\ref{lem:univ001}. This is a 
    contradiction to our assumption that $\bdiff{\univ_w(s)} = k$ for all $s 
    \in \N$. Thus there cannot be such an $s_0$.\qed
\end{proof}

%% file: growth.tex
In this section, our goal is to find a characterisation for the previous example:
for $w=\tb\ta\tb\tc\tc\ta\tb\tc$, we 
have $k = \univ(w) = 2$. Then we obtain $\univ(w^2)=2k+1$ and $\univ(w^3)=3k+1$ 
again. This means $\bdiff{\univ_w(2)} = k+1$, but $\bdiff{\univ_w(3)} = 
k$. Thus, we seek 
criteria to determine whether $\bdiff{\univ_w(s_0)}$ is $k$ or $k+1$ for a 
given $s_0$. Having this purpose in mind, we now look at the arch factorisation of 
$w^2$: the 
first $k$ arches of $w^2$ are those of $w$ but the next arch begins with 
the remainder $\rem(w)$ (which may be empty) and ends with a non-empty prefix $p$ of 
$w$. The 
remaining arches are those of $p^{-1}w$. Thus $\univ(w^2) = k+1+ 
\univ(p^{-1}w)$ or, equivalently, $\bdiff{\univ_w(2)} = 1 + \univ(p^{-1}w)$.  
Clearly, by increasing the length of $p$, we decrease $\univ(p^{-1}w)$. So we 
get $\bdiff{\univ_w(2)} = k+1$ iff $p$ is not {\em too long}.  Thus, how long does $p$ have to be, 
and what is the longest prefix $p$ that we can remove from $w$ without reducing its universality? 
Answering these questions will yield the 
characterisation that we are striving to. Regarding the first question, $p$ must be long enough for 
the equation $\alf(\rem(w)p) = \alphabet$ 
to hold, since only then $\rem(w)p$ is an arch of $w^2$. The second question is 
answered by the following lemma.

\ifpaper
\begin{lemma}[$\ast$]
\else
\begin{lemma}
\fi
    \label{univ:lem:prefix}
    For $w\in\alphs$, the word $p=\rem(w^R)^R$ is the 
    longest prefix of $w$ such that $\univ(p^{-1}w) = \univ(w)$ holds.
\end{lemma}
\ifpaper
\else
\input{proof_prefix}
\fi

For instance, if $w=\mathtt{nabananab}$, we have $\rem(w^R)^R=\mathtt{an}$, and indeed,
if we remove $\mathtt{na}$ from the beginning of $w$, $\iota(\mathtt{bananab})=2$ still holds.

In our considerations above we looked at the arch factorisation of $w^2$. This 
can be generalised to the concatenation of 
two arbitrary words $w$ and $u$. The answer to the question, whether 
$\univ(wu)$ equals $\univ(w) + \univ(u) + 1$ or $\univ(w) + \univ(u)$, is answered by a
slightly more general version of 
\cite[Proposition~18]{barker2020scattered}.

\ifpaper
\begin{proposition}[$\ast$]
\else
\begin{proposition}
\fi
    \label{univ:prop:chargrowth}
    Let $u,w\in\alphs$ with $k=\univ(w)$ and $\ell=\univ(u)$. Then $\univ(wu) 
    = k + \ell + 1$ if and only if $\alf(\rem(w)\rem(u^R)) = \alphabet$.
\end{proposition}
\ifpaper
\else
\input{proof_chargrowth}
\fi

Proposition~\ref{univ:prop:chargrowth} implies immediately the desired characterisation of the 
growth of
$\univ_w$, i.e. $\bdiff{\univ_w}$, using $w^{s} = w^{s-1} \cdot w$.

\ifpaper
\begin{corollary}[$\ast$]
\else 
\begin{corollary}
\fi
    \label{univ:cor:chardiffrem}
    Let $w \in \alphs$, $k=\univ(w)$ and $s\in\N$. Then we have 
    $\bdiff{\univ_w(s)} = k+1$ if and only if $\alf(\rem(w^{s-1})\rem(w^R)) = 
    \alphabet$.
\end{corollary}
\ifpaper
\else
\input{proof_chardiffrem}
\fi

Corollary~\ref{univ:cor:chardiffrem} is fundamental to the remainder of 
this chapter. It gives us a useful tool to investigate the universality of 
repetitions $w^s$. It implies that the growth $\bdiff{\univ_w}$ and the 
remainder mapping $s \mapsto \rem(w^s)$ of repetitions depend on each other. 
Thus we can gain insight into the behaviour of $\bdiff{\univ_w}$ by studying 
$s \mapsto \rem(w^s)$ and vice versa. Especially, we are interested in question under which 
circumstances we gain eventual periodicity in the growth.
Notice, that $\rem(w^s)$ depends recursively on $\rem(w^{s-1})$ by the equation 
$\rem(w^s) = \rem(w^{s-1} \cdot w)$. However, we can slightly refine this notion:
since removing whole arches from a word does not change its remainder, we have the 
following observation.

\begin{remark}
    Let $w\in\alphs$ and $s \in \N$. Then $\rem(w^s) = \rem(\rem(w^{s-1}) w)$.
Thus, there is always a word 
$u \in \alphs$ with $\alf(u) \subsetneq \alphabet$ such that $\rem(w^s) = \rem(uw)$.  
\end{remark}

Now, we investigate when $\rem(uw)$ and $\rem(w)$ differ (or more general $\rem(uw)$
and $\rem(vw)$ for some $v \in \alphs\backslash\{u\}$ with 
$\alf(v) \subsetneq \alphabet$). Consider for motivation, $w = (\ta\tb\tb\tc) \cdot 
(\tc\ta\ta\tb) \cdot \tc\ta\ta$. If $u\in\{\ta,\tb\}^{\ast}$ then in $uw$,
$u$ simply adds to the first arch of $w$ without bringing any 
significant change into the arch factorisation. This happens because 
the last letter of an arch is unique and $\tc$ does not occur in $u$ leading to
$\rem(uw) = \rem(w)$. Now consider $u = \tc$, This time, regarding $uw$, $u$ causes the factor 
$\tb\tc$ to get \emph{released} from $w$'s first arch and $\ta\tb$ gets released from 
$w$'s second arch and \emph{binds} $\tc$, the first letter of $w$'s remainder, 
to build a third arch. Thus by prepending $u$ to $w$, changes in the arch 
factorisation get carried through the whole word. Finally, let $u = \tc\ta\ta$.  The arch 
factorisations of $\tc w$ and $uw$ do not differ 
significantly from each other since the last letter of $\arch_1(\ta 
w)$ is $\tb$ and adding any number of the letter $\ta$ to $\tc$ has no effect.
Notice,  that in all cases, we only argued by $\alf(u)$ but neither on the number of letters nor 
their position. In fact, for any $u \in \alphs$ with $\alf(u) \subsetneq \alphabet$, $\alf(u)$ is 
the only relevant information about $u$ for the computation of 
$\rem(uw)$, since we assumed $\iota(w)>0$.

\ifpaper
\begin{lemma}[$\ast$]
\else 
\begin{lemma}
\fi
    \label{univ:lem:d}
    If $u,v,w \in \alphs$ with $\alf(u) = \alf(v) \subsetneq \alphabet$ then $\rem(u w) 
= \rem(v w)$ holds.
\end{lemma}

\ifpaper
\else
\input{proof_d}
\fi

Now, we return our attention to the equation $\rem(w^{s_0}) 
= \rem(\rem(w^{s_0-1}) w)$ for some $s_0\in\N_0$. If the letters occurring in $\rem(w^{s_0})$ 
and $\rem(w^{s_0-1})$ are the same, Lemma~\ref{univ:lem:d} implies that $s 
\mapsto \rem(w^s)$ stays constant beginning at $s_0$. And even more general, we 
get in the same way that, if for some $t_0\in\N_0$ the letters occurring in 
$\rem(w^{s_0})$ and $\rem(w^{t_0})$ are the same, then $\rem(w^{s_0+1}) = 
\rem(w^{t_0+1})$, i.e. the mapping $s \mapsto \rem(w^s)$ is periodic 
beginning at $\min\set{s_0, t_0}$. The following lemma proves this observation.  
Recall that we defined $\alfrem{w} = \alf(\rem(w))$. 

\ifpaper
\begin{lemma}[$\ast$]
\else
\begin{lemma}
\fi
    \label{univ:lem:eventperiodrem}
    Let $w\in\alphs$ and let $s,t\in\N_0$. If 
    $\alfrem{w^s} = \alfrem{w^t}$ then $\rem(w^{s+i}) = \rem(w^{t+i})$ 
    for all $i\in\N$.
\end{lemma}
\ifpaper
\else
\input{proof_eventperiodrem}
\fi

The converse of Lemma~\ref{univ:lem:eventperiodrem} is not necessarily 
 true:  considering $w = (\ta\tc\ta\tb) \cdot \tb$, we have $\rem(w^2) = \rem(w^3) = \rem(w^s)$ for 
all $s 
\geq 2$, but 
    $\alfrem{w} \neq \alfrem{w^2}$. Notice also, that  
Lemma~\ref{univ:lem:eventperiodrem} does not hold for $i=0$. For $w = (\ta\tb) \cdot \tb$  we have 
$\alfrem{w} = \alfrem{w^2}$, but $\rem(w) \neq 
    \rem(w^2)$. But indeed, if we only 
care about the letters in the remainders, then we can extend 
Lemma~\ref{univ:lem:eventperiodrem} to all $i \in \N_0$ and  the converse 
immediately holds as well.

\ifpaper
\begin{lemma}[$\ast$]
\else
\begin{lemma}
\fi
    \label{univ:prop:eventperiodalfrem}
    Let $w\in\alphs$ and let $s,t\in\N_0$. Then we have 
    $\alfrem{w^{s+i}} = \alfrem{w^{t+i}}$ for all $i\in\N_0$ if and only if 
    $\alfrem{w^s} = \alfrem{w^t}$.
\end{lemma}
\ifpaper
\else
\input{proof_eventperiodalfrem}
\fi

Lemma~\ref{univ:prop:eventperiodalfrem} implies the desired periodicity 
property of the growth.

\ifpaper
\begin{proposition}[$\ast$]
\else
\begin{proposition}
\fi
\label{evperiodic}
The growth of the universality index, $\nabla\iota_w$, is eventually periodic.
\end{proposition}
\ifpaper
\else
\input{proof_evperiodic}
\fi

The 
following lemma shows that $s$ and $t$ can be bound by $\sigma$ and leads
to a theorem capturing the above considerations.

\ifpaper
\begin{lemma}[$\ast$]
\else
\begin{lemma}
\fi
    \label{univ:lem:boundalfrems}
    For all $w \in \alphs$ we have $\abs{\set{\alfrem{w^s} | s \in\N_0}} \leq 
    \asize$.
\end{lemma}
\ifpaper
\else 
\input{proof_boundalfrems}
\fi

\ifpaper
\begin{theorem}[$\ast$]
\else 
\begin{theorem}
\fi
    \label{univ:theo:periodicity}
    For all $w \in \alphs$ there exist 
    $s,t\in\disinv{0}{\asize}$ with $s<t$ such that\\
    1. $\rem(w^{s+i}) = \rem(w^{t+i})$ for all $i \in \N$,\\
    2. $\alfrem{w^{s+i}} = \alfrem{w^{t+i}}$ for all $i \in \N_0$, and\\
    3. $\bdiff{\univ_w(s+i)} = \bdiff{\univ_w(t+i)}$ for all $i \in \N$.
\end{theorem}
\ifpaper
\else
\input{proof_periodicity}
\fi

Theorem~\ref{univ:theo:periodicity} states that beginning at $s+1$, 
$\bdiff{\univ_w}$ is periodic of length $t-s$. So, given an 
$n\in\N$ we can divide $\disinv{1}{n}$ into $\disinv{1}{s}$ (before $\bdiff{\univ_w}$ is periodic) 
and $\disinv{s+1}{n}$ (containing the periodic part). By division with remainder, we obtain
$\ell,m\in\N_0$ such that $n-s = \ell \cdot (t-s) + m$ and thus, the subinterval 
$\disinv{s+1}{s+\ell(t-s)}$ contains $\ell$ full periods while $\disinv{s+\ell(t-s)+1}{n}$ is the 
rest. This observation motivates the 
following Proposition.

\ifpaper
\begin{proposition}[$\ast$]
\else 
\begin{proposition}
\fi\label{division}
    Let $w \in \alphs$ and let $n \in 
    \N$ with $n \geq \asize$. Then there exist $\ell,m,s,t\in\N_0$ with  
    $s<t\leq\asize$, $m < t-s$, and $n-s = \ell \cdot (t-s) + m$ 
    such that $\univ_w(n) = \univ_w(s+m) + \ell \cdot (\univ_w(t) - 
    \univ_w(s))$.
\end{proposition}
\ifpaper
\else 
\input{proof_division}
\fi

\ifpaper
\begin{proposition}[$\ast$]
\else
\begin{proposition}
\fi\label{consttime}
Given $w\in\Sigma^{\ast}$, we can compute $\univ_w(n)$ for all $n\in\N_0$ in constant time with a 
preprocessing time of $\bigo(\sigma|w|)$.
\end{proposition}
\ifpaper
\else
\input{proof_consttime}
\fi

In the rest of this work we will usually only need the notion of constancy 
and not of periodicity. Thus we restate
Lemma~\ref{univ:lem:eventperiodrem} and 
Proposition~\ref{univ:prop:eventperiodalfrem} for the case $s = t-1$. The claims 
follow directly by the lemma and the proposition.

\begin{lemma}
    \label{univ:lem:e}
    Let $w\in\alphs$ and $s_0\in\N$. If 
    $\alfrem{w^{s_0-1}} = \alfrem{w^{s_0}}$ then $\rem(w^s) = \rem(w^{s_0})$ 
    for all $s \geq s_0$.
\end{lemma}

\begin{proposition}
    \label{univ:prop:constant}
    Let $w\in\alphs$and $s_0\in\N$. Then we have 
    $\alfrem{w^s} = \alfrem{w^{s_0}}$ for all $s\geq s_0-1$ if and only if 
    $\alfrem{w^{s_0-1}} = \alfrem{w^{s_0}}$.
\end{proposition}

Hence, if $\alfrem{w^{s_0-1}} = \alfrem{w^{s_0}}$ holds for any $s_0\in\N$ then 
    the mappings $s \mapsto \rem(w^s)$ and $s \mapsto \alfrem{w^s}$ are both 
    eventually constant.

%% file: proof_prefix.tex
\begin{proof}
    By reversing the arch factorisation of $w^R$ we can write $w$ as
    \[
    w = \rem(w^R)^R \cdot \prod_{i=1}^k \arch_{k+1-i}(w^R)^R.
    \]
    There one can see that $\univ(p^{-1} w) = k$ still holds. Next we have to 
    show that the universality becomes smaller when taking a longer prefix, 
    i.e. $\univ(q^{-1}w) < k$ for any prefix $q$ of $w$ with $\abs{q} > 
    \abs{p}$. It 
    suffices to show this for the prefix $q$ with length $\abs{p}+1$.
    Let $\tc$ be the first letter of the arch $\arch_k(w^R)^R$ and $y_k \in 
    \alphs$ the suffix such that
    \[
    \arch_k(w^R)^R = \tc y_k.
    \]
    Then $q=p\tc$ is the prefix of $w$ of length $\abs{p} + 1$. Now we show 
    that $\univ(q^{-1} w) = k-1$ by building its arch factorisation.  
    Since $\tc$ is the last letter of $\arch_k(w^R)$ it is unique in this 
    arch. Thus, the letter $\tc$ does not occur in $y_k$, which implies that 
    $\univ(y_k) = 0$. If now $w$ had only one arch then
    \[
    \univ(q^{-1}w) = \univ(y_k) = 0 < k.
    \]
    would follow and thus our claim holds. So let us assume that $k>1$.  Then 
    $y_k$ still needs a non-empty prefix of $\arch_{k-1}(w^R)^R$ as suffix to 
    build a full arch. Hence, there exist $x_{k-1}\in\alphp$ and 
    $y_{k-1}\in\alphs$ such that
    \[\begin{aligned}
        \arch_{k-1}(w^R)^R &= x_{k-1} y_{k-1}, \\
        \arch_1(q^{-1}w) &= y_k x_{k-1}.
    \end{aligned}\]
    But then again it follows that $\univ(y_{k-1})=0$. Consequently, by 
    iterating this process of building the arches of $q^{-1}w$ we get for all 
    $i \in \disinv{1}{k-1}$ some factors $x_{k-i}\in\alphp$ and 
    $y_{k-i}\in\alphs$ such that
    \[\begin{aligned}
        \arch_{k-i}(w^R)^R &= x_{k-i} y_{k-i}, \\
        \arch_i(q^{-1}w) &= y_{k-i+1} x_{k-i}, \\
        \univ(y_{k-i}) &= 0.
    \end{aligned}\]
    It follows that first $\rem(q^{-1}w) = y_1$ and second $q^{-1}w$ has 
    exactly $k-1$ arches, and hence $\univ(q^{-1}w) = k-1$, which was to be 
    shown.\qed
\end{proof}

%% file: proof_chargrowth.tex
\begin{proof}
    From the arch factorisations of $w$ and $u^R$ respectively follows 
    immediately that if
    \begin{equation}
        \alf(\rem(w)\rem(u^R)) = \alphabet
    \end{equation}
    then $\univ(wu) = k + \ell + 1$.
    
    Now on the other hand assume that
    \begin{equation}
        \label{eq:univ:prop:chargrowth:1}
        \alf(\rem(w)\rem(u^R))\neq\alphabet.
    \end{equation}
    We can factorise $wu$ as
    \[
    wu = w(u^R)^R =
    \left[\prod_{i=1}^k \arch_i(w)\right] \cdot \rem(w) 
    \cdot
    \rem(u^R)^R \cdot \left[\prod_{i=1}^\ell 
    \arch_{\ell+1-i}(u^R)^R\right],
    \]
    i.e. as the product of the arch factorisation of $w$ and the reversal of 
    the arch factorisation of $u^R$. Then one can see that
    \[
    \univ(wu) = k + \univ(\rem(w)u).
    \]
    Now it is left to show that $\univ(\rem(w)u) < \ell + 1$. We will 
    determine $\univ(\rem(w)u)$ by looking as usual at its arch factorisation. 
    First, there exists a prefix $p$ of $u$ such that the first arch of 
    $\rem(w)u$ is
    \[
    \arch_1(\rem(w)u) = \rem(w) p.
    \]
    Then it follows from Equation \eqref{eq:univ:prop:chargrowth:1} that $p$ has to 
be 
    longer than $\rem(u^R)^R$, i.e. $\abs{p} > \abs{\rem(u^R)^R}$. And so, 
    Lemma~\ref{univ:lem:prefix} tells us that $\univ(p^{-1}u) < \ell$ holds.  
    Therefore, finally,
    \[
    \univ(\rem(w)u) = \univ(\rem(w)p) + \univ(p^{-1}u) < 1 + \ell.\qed
    \]
\end{proof}

%% file: proof_chardiffrem.tex
\begin{proof}
    The claim follows by Proposition~\ref{univ:prop:chargrowth} 
    applied on on $w^{s-1}$ and $w$. \qed
\end{proof}

%% file: proof_d.tex
\begin{proof}
    The first arch of $uw$ is formed by $u$ and the shortest prefix $p$ of $w$ 
    such that $\alf(u) \cup \alf(p) = \alphabet$. Thus $p$ depends solely on 
    $\alf(u)$. This makes $p$ also the shortest prefix such that $\alf(v) 
    \cup \alf(p) = \alphabet$. Hence
    \[
    \rem(uw) = \rem(p^{-1}w) = \rem(vw).\qed
    \]
\end{proof}

%% file: proof_eventperiodrem.tex
\begin{proof}
    Let $\alfrem{w^s} = \alfrem{w^t}$. We show by induction that 
    $\rem(w^{s+i}) = \rem(w^{t+i})$ for all $i\in\N$. For $i=1$, we
    get by Lemma~\ref{univ:lem:d} that
    \[
        \rem(w^{s+1}) = \rem(\rem(w^s) w) = \rem(\rem(w^t) w) =  \rem(w^{t+1}).
    \]
    For $i>1$, assume that $\rem(w^{s+i}) = \rem(w^{t+i})$. Then we get
    \[
    \rem(w^{s+i+1}) = \rem(\rem(w^{s+i}) w) = \rem(\rem(w^{t+i}) w) = 
    \rem(w^{t+i+1}). \qed
    \]
\end{proof}

%% file: proof_eventperiodalfrem.tex
\begin{proof}
    Let $\alfrem{w^s} = \alfrem{w^t}$. Then 
    Lemma~\ref{univ:lem:eventperiodrem} implies that $\rem(w^{s+i}) = 
    \rem(w^{t+i})$ for all $i\in\N$. Hence $\alfrem{w^{s+i}} = \alfrem{w^{t+i}}$ 
    for all $i \in \N_0$.
    
    The other direction follows immediately by $i=0$.\qed
\end{proof}

%% file: proof_evperiodic.tex
\begin{proof}
Since the mapping $s \mapsto \alfrem{w^s}$ can only have finitely many 
values ($\Sigma$ is finite), there exist $s_0,t_0\in\N_0$ with $s_0 \neq t_0$ 
such that $\alfrem{w^{s_0}} = \alfrem{w^{t_0}}$ and thus 
Lemma~\ref{univ:prop:eventperiodalfrem} implies that $s \mapsto 
\alfrem{w^s}$ is periodic.
The claim follows by Corollary~\ref{univ:cor:chardiffrem}.
\qed
\end{proof}

%% file: proof_boundalfrems.tex
\begin{proof}
    For all $s \in \N_0$ the remainder $\rem(w^s)$ is a suffix of $w$. 
    Therefore these remainders are pairwise suffix-compatible with each other, 
    and consequently also the sets $\alfrem{w^s}$ are pairwise comparable 
    regarding inclusion, i.e. the set $\set{\alfrem{w^s} | s \in \N_0}$ is totally 
ordered. Thus we can order them by some bijective mapping $\pi 
    : \N_0 \rightarrow N_0$ such that
    \[
    \alfrem{w^{\pi(0)}} \supseteq \alfrem{w^{\pi(1)}} \supseteq \ldots.
    \]
    But since also each set contains less than $\asize$ elements, i.e. $0 \leq 
    \abs{\alfrem{w^s}} < \asize$ for all $s \in \N_0$, it follows that $s 
    \mapsto \alfrem{w^s}$ has at most $\asize$ different values.\qed
\end{proof}

%% file: proof_periodicity.tex
\begin{proof}
    By Lemma~\ref{univ:lem:boundalfrems} there exist 
    $s<t\in\disinv{0}{\asize}$ such that $\alfrem{w^s} = \alfrem{w^t}$. Then 
    the first claim follows by Lemma~\ref{univ:lem:eventperiodrem} and the 
    second by Proposition~\ref{univ:prop:eventperiodalfrem}. The third claim 
    follows by combining the second claim with 
    Corollary~\ref{univ:cor:chardiffrem}.\qed
\end{proof}

%% file: proof_division.tex
\begin{proof}
    Let $s$ and $t$ be minimal such that $s<t$ and $\alfrem{w^s} = 
    \alfrem{w^t}$. By Theorem~\ref{univ:theo:periodicity} we have $s,t \leq 
    \asize$.
    Then there are by division with remainder $m,\ell\in\N_0$ such that $n-s = 
    \ell \cdot (t-s) + m$ with $m < t-s$. It follows by 
    Theorem~\ref{univ:theo:periodicity} that
    \[\begin{aligned}
        \univ_w(n) &= \sum_{i=1}^{n} \bdiff{\univ_w(i)}\\
        &= \left[\sum_{i=1}^{s} \bdiff{\univ_w(i)} \right] + \ell \cdot 
        \left[\sum_{i=s+1}^{t} \bdiff{\univ_w(i)} \right] + 
        \left[\sum_{i=s+1}^{s+m} \bdiff{\univ_w(i)} \right]  \\
        &= \univ_w(s+m) + \ell \cdot \left[\sum_{i=s+1}^{t} 
        \bdiff{\univ_w(i)} \right] \\
        &= \univ_w(s+m) + \ell \cdot (\univ_w(t) - \univ_w(s)),
    \end{aligned}\]
    which was to be shown.\qed
\end{proof}

%% file: proof_consttime.tex
\begin{proof}
    We only have to compute the minimal $s,t\in\N_0$ with $\alfrem{w^s} = \alfrem{w^t}$ and 
$\univ_w(i)$ for all $i 
    \in \disinv{0}{\asize}$ once, taking in total $\bigo(\sigma|w|)$ time. Afterwards, by 
Proposition~\ref{division}, we can compute $\univ_w(n)$ for all $n \in \N_0$ in constant time.\qed
\end{proof}

%% file: chain.tex
Section~\ref{univ:sec:powrem} established by 
Corollary~\ref{univ:cor:chardiffrem} a correspondence between the growth 
$\bdiff{\univ_w}$ and the remainder mapping $s \mapsto \rem(w^s)$. We achieved 
this corollary by interpreting $w^{s}$ recursively as $w^{s-1} \cdot w$. But 
if we interpret it as $w \cdot w^{s-1}$ instead then we find another useful 
relationship that we capture in the following lemma.

\ifpaper
\begin{lemma}[$\ast$]
\else 
\begin{lemma}
\fi
    \label{univ:lem:remsuf}
    Let $w \in \alphs$ and $s\in\N_0$. Then the following 
    two statements hold:\\
    1. If $\bdiff{\univ_w(s+1)} = \univ(w)$ then $\rem(w^s)$ is a suffix of 
        $\rem(w^{s+1})$. \\
    2. If $\bdiff{\univ_w(s+1)} = \univ(w)+1$ then $\rem(w^{s+1})$ is a suffix 
        of $\rem(w^s)$.
\end{lemma}
\ifpaper
\else
\input{proof_remsuf}
\fi

Combining Corollary~\ref{univ:cor:chardiffrem} and Lemma~\ref{univ:lem:remsuf} 
implies that if $\rem(w^s)$ is {\em long enough} then $\bdiff{\univ_w(s+1)} = k+1$, which 
implies that $\rem(w^{s+1})$ is a suffix of $\rem(w^s)$. Thus, now 
$\rem(w^{s+1})$ may have become {\em so short} that in the next step we get 
$\bdiff{\univ_w(s+2)} = k$. This is exactly what happens for $w = 
(\mathtt{babc})\cdot(\mathtt{caab})\cdot\tc$.
The other case is symmetrical: if $\rem(w^s)$ is not {\em long enough} then 
$\bdiff{\univ_w(s+1)} = k$, which implies that $\rem(w^s)$ is a suffix of 
$\rem(w^{s+1})$. Thus, now $\rem(w^{s+1})$ may have become {\em long enough} for 
$\bdiff{\univ_w(s+2)} = k+1$. Also note, that the converses of 
Lemma~\ref{univ:lem:remsuf} do not necessarily 
hold if $\rem(w^s) = \rem(w^{s+1})$. Considering  $w = (\ta\tb)\cdot\ta$, we get 
$\bdiff{\univ_w(2)} = 1$ 
and $\rem(w) 
        = \ta = \rem(w^2)$. On the other hand, if $w = (\ta\ta\tb)\cdot\tb$  then $\univ(w) 
= 1$, but $\bdiff{\univ_w(2)} = 2$ and $\rem(w) = \tb = \rem(w^2)$.
But as soon as two successive remainders are equal or even if only their set of occurring 
letters are equal, then $s \mapsto 
\rem(w^s)$ is eventually constant. Thus, it 
seems useful to consider the cases, where these conditions are excluded, 
explicitly. We state them in the following two corollaries.

\begin{corollary}
    \label{univ:cor:sufremoneone}
    Let $w\in\alphs$ and $s\in\N_0$. If $\rem(w^s) \neq \rem(w^{s+1})$, then\\
    1. $\bdiff{\univ_w(s+1)} = k$ iff $\rem(w^s)$ is a 
        (proper) suffix of $\rem(w^{s+1})$ and \\
    2. $\bdiff{\univ_w(s+1)} = k+1$ iff $\rem(w^{s+1})$ is a 
        (proper) suffix of $\rem(w^s)$.
\end{corollary}

\begin{corollary}
    \label{univ:cor:sufremalphchain}
    Let $w\in\alphs$ and $s\in\N_0$ with $\alfrem{w^s} \neq 
    \alfrem{w^{s+1}}$. Then we obtain\\
    1. $\bdiff{\univ_w(s+1)} = k$  iff $\rem(w^{s})$ is a (proper) suffix of 
$\rem(w^{s+1})$  iff $\alfrem{w^s} \subsetneq \alfrem{w^{s+1}}$ and
2. $\bdiff{\univ_w(s+1)} = k+1$ iff $\rem(w^{s+1})$ is a (proper) suffix of 
$\rem(w^{s})$  iff $\alfrem{w^{s}} \supsetneq \alfrem{w^{s+1}}$.
\end{corollary}

However, usually we do not know whether $s \mapsto \rem(w^s)$ is eventually 
constant, i.e. whether Corollary~\ref{univ:cor:sufremalphchain} is applicable.  
So now the following lemma gives a criterion to decide whether this condition 
is satisfied.

\ifpaper
\begin{lemma}[$\ast$]
\else 
\begin{lemma}
\fi
    \label{univ:lem:eventconstcrit}
    For all $w \in \alphs$,  $s \mapsto 
    \rem(w^s)$ is eventually constant iff $\bdiff{\univ_w}$ is.
\end{lemma}
\ifpaper
\else
\input{proof_eventconstrict}
\fi

Applying Lemma~\ref{univ:lem:eventconstcrit} gives a new insight regarding the 
characterisation when $\circuniv(w)=\iota(w)+1$ holds.

\begin{corollary}
    \label{univ:cor:smallappchain}
    Let $w \in \alphs$ and $k=\univ(w)$. If $\circuniv(w) = k+1$ then $s 
    \mapsto \rem(w^s)$ is eventually constant.
\end{corollary}

We will usually apply Lemma~\ref{univ:lem:remsuf} and Corollary~\ref{univ:cor:sufremalphchain}
in the following way: an interval $\disinv{\ell+1}{n}$ on 
which we have $\bdiff{\univ_w(s)} = k$ implies that there is an ascending chain
$\alfrem{w^\ell} \subseteq \alfrem{w^{\ell+1}} \subseteq \ldots \subseteq 
\alfrem{w^n}$. But notice that we cannot 
follow that $\bdiff{\univ_w(s)} = k$ holds on $\disinv{\ell+1}{n}$ by the existence of such a chain 
since there may be 
equality in some steps.
However, if we exclude $\alfrem{w^s} = \alfrem{w^n}$ for 
all $s \geq n-1$, i.e. $s \mapsto \alfrem{w^s}$ is not yet 
constant then we know that the chain from $\alfrem{w^\ell}$ to $\alfrem{w^n}$
is strict, and such a strictly ascending chain implies $\bdiff{\univ_w(s)} = k$ 
on $\disinv{\ell+1}{n}$ (the case where $\bdiff{\univ_w(s)} = k+1$ on 
$\disinv{\ell+1}{n}$ is symmetrical with descending chains). This way Lemma~\ref{univ:lem:remsuf} 
and 
Corollary~\ref{univ:cor:sufremalphchain} can be used to translate questions 
about $\univ_w$ and $\circuniv_w$ into questions about chains of sets. 

In the following two subsections we investigate ascending and descending chains 
in more detail. By improvements of Lemma~\ref{univ:lem:eventconstcrit}, we are 
able to generalise the two aforementioned results from \cite{barker2020scattered}.

\bigskip

\noindent
\textbf{4.1 Ascending Chains.}\quad
So far, we established that an interval 
$\disinv{\ell+1}{n}$ on which $\bdiff{\univ(w^s)} = k$ holds implies an 
ascending chain $\alfrem{w^\ell} \subseteq \ldots \subseteq \alfrem{w^n}$ and 
if that chain is strict then the implication holds also in the other direction. 
The following proposition gives us a structural property about a strictly 
ascending chain 
of 
length exactly $\asize$, where $\asize$ is the size of the alphabet.

\ifpaper
\begin{lemma}[$\ast$]
\else 
\begin{lemma}
\fi
    \label{univ:lem:h}
    Let $w \in \alphs$ and $\ell \in \N_0$. If
    $\alfrem{w^\ell} \subsetneq \ldots \subsetneq \alfrem{w^{\ell+\asize-1}}$
    is a strictly ascending chain of length $\asize$ then we have 
$\abs{\alfrem{w^s}} 
    = s-\ell$ for all $s \in \disinv{\ell}{\ell + \asize - 1}$.
\end{lemma}
\ifpaper
\else 
\input{proof_h}
\fi

By Lemma~\ref{univ:lem:h}, strictly ascending chains of length $\sigma+1$ 
cannot exist.

\ifpaper
\begin{corollary}[$\ast$]
\else
\begin{corollary}
\fi
    \label{univ:cor:longestascendingchain}
    Let $w \in \alphs$, $\ell,n \in \N_0$, and $\alfrem{w^\ell} \subsetneq \ldots 
\subsetneq \alfrem{w^{\ell+n-1}}$  be a strictly ascending chain of length $n$. 
Then $n \leq \asize$.
\end{corollary}
\ifpaper
\else
\input{proof_longestachain}
\fi

\begin{remark}
In fact, there actually exists a strictly ascending 
chain of length $\asize$. For $\alphabet = \set{\ta_1, \dots, \ta_\asize}$ set
$w = \prod_{i=1}^\asize \ta_i^2$. Then we have $\univ(w) = 1$, $\bdiff{\univ_w(s)} = 
1$ for all $s \in \disinv{1}{\asize-1}$ and $\bdiff{\univ_w(\asize)} = 
2$. Furthermore $\alfrem{w^0} \subsetneq \ldots \subsetneq \alfrem{w^{\asize-1}}$
    is a strictly ascending chain of length $\asize$.
\end{remark}

Corollary~\ref{univ:cor:longestascendingchain} leads to the following 
proposition, which states that if we have $\bdiff{\univ_w(s)} = k$ for the first 
$\asize-1$ repetitions then $\bdiff{\univ_w}$ is already constant. 

\ifpaper
\begin{proposition}[$\ast$]
\else 
\begin{proposition}
\fi
    \label{univ:prop:k}
    Let $w \in \alphs$, $k=\univ(w)$. If 
    $\bdiff{\univ_w(s)} = k$ for all $s \in \disinv{1}{\asize}$ then 
    $\bdiff{\univ_w(s)} = k$ for all $s \in \N$.
\end{proposition}
\ifpaper
\else 
\input{proof_k}
\fi

Even though the bound $\asize$ in Proposition~\ref{univ:prop:k} is tight, we can 
still improve the statement 
in another way. If we consider the \emph{circular} universality $\circuniv_w$ 
instead of the plain universality $\univ_w$ then we can lower the bound to 
$\asize-1$. Before we present the corresponding proposition, we prove two auxiliary 
lemmata.

\ifpaper
\begin{lemma}[$\ast$]
\else
\begin{lemma}
\fi
    \label{univ:lem:l}
    Let $w \in \alphs$ and $k = \univ(w)$. If each letter $\ta \in \alphabet$ 
    occurs only once in each arch of $w$ and at most once in the remainder 
    $\rem(w)$ then $\bdiff{\circuniv_w(s)} = k$ for all $s \in \N$.
\end{lemma}
\ifpaper
\else 
\input{proof_1}
\fi

\ifpaper
\begin{lemma}[$\ast$]
\else
\begin{lemma}
\fi
    \label{univ:lem:m}
    Let $w \in \alphs$, $k=\univ(w)$. Let 
    there be a word $y\in\alphs$ and a letter $\ta\in\alphabet$ such that 
    $\abs{\alf(y)} \leq \asize-2$ and such that $\ta y \ta$ is a factor of 
    some conjugate $v$ of $w$. If $\bdiff{\circuniv_w(s)} = k$ holds for all 
    $s \in \disinv{1}{\asize-1}$ then we have $\bdiff{\univ_w(s)} = k$ for 
    all $s \in \N$.
\end{lemma}
\ifpaper
\else 
\input{proof_m}
\fi

\ifpaper
\begin{proposition}[$\ast$]
\else
\begin{proposition}
\fi
    \label{univ:theo:n}
    Let $w \in \alphs$. Then we have
    $\bdiff{\univ_w(s)} = k$ for all $s \in \N$ if and only if 
    $\bdiff{\circuniv_w(s)} = k$ for all $s \in \disinv{1}{\asize-1}$.
\end{proposition}
\ifpaper
\else
\input{proof_n}
\fi

With Proposition~\ref{univ:theo:n} we can finally achieve our first main goal of 
generalising Theorem~\ref{theo:barker23} to alphabets of arbitrary size. 
\begin{theorem}
    Let $w \in \alphs$ with $k=\univ(w)>0$ and let $s \in \N$. \\
    1. If $\circuniv(w) = k+1$ then $\univ(w^s) = sk+s-1$.\\
    2. If $\bdiff{\circuniv_w(t)} = k$ for all $t \in 
        \disinv{1}{\asize-1}$ then $\univ(w^s) = sk$.
\end{theorem}
\begin{proof}
    The claim follows by combining Theorem~\ref{theo:barker22} and 
    Proposition~\ref{univ:theo:n}.\qed
\end{proof}

\begin{remark}
Considering again $w = \prod_{i=1}^\asize \ta_i^2$ shows that the bound $\asize-1$ 
in Proposition~\ref{univ:theo:n} is tight. The word $u = \ta_1 \left[ 
\prod_{i=2}^n \ta_i^2 \right] \ta_1$ is a conjugate of $w$ with 
$\bdiff{\univ_u(s)} = 1$ for all $s \leq	\asize-2$ and 
$\bdiff{\univ_u(\asize-1)} = 2$. Thus we have
    $\bdiff{\circuniv_w(s)} = 1$ for all $s \leq \asize-2$ and 
    $\bdiff{\circuniv_w(\asize-1)} = 2$.
 \end{remark}


\bigskip

\noindent
\textbf{4.2 Descending Chains.}\quad
Now, we discuss descending chains instead of ascending chains. We 
begin by searching for the longest strictly descending chain that is possible. 
The following proposition gives us a structural property about such a chain of 
length exactly $\asize$ and is symmetrical to Lemma~\ref{univ:lem:h}.

\ifpaper
\begin{lemma}[$\ast$]
\else 
\begin{lemma}
\fi
    \label{univ:lem:i}
    Let $w \in \alphs$ and $\ell \in \N_0$. If $\alfrem{w^\ell} \supsetneq \ldots 
\supsetneq \alfrem{w^{\ell+\asize-1}}$ is a strictly descending chain of length 
$\asize$ then 
    $\abs{\alfrem{w^s}} = \asize + \ell - s$ for all $s \in \disinv{\ell + 
    1}{\ell + \asize}$.
\end{lemma}
\ifpaper
\else
\input{proof_i}
\fi

Analogously to Corollary~\ref{univ:cor:longestascendingchain}, there cannot be a strictly descending 
chain of length $\asize+1$.  Surprisingly, such a chain of length $\asize$ leads to a 
contradiction as well. First, we present an auxiliary lemma.

\ifpaper
\begin{lemma}[$\ast$]
\else
\begin{lemma}
\fi
    \label{univ:lem:wlogellequalsone}
    Let $w \in \alphs$ and $\ell,n \in \N_0$. If $\alfrem{w^\ell} \supsetneq 
\ldots \supsetneq \alfrem{w^{\ell+\asize-1}}$ is a strictly descending chain of 
length $\asize$ then $\alfrem{w} \supsetneq \ldots \supsetneq \alfrem{w^{\asize}}$
    is one as well.
\end{lemma}
\ifpaper
\else
\input{proof_equalsone}
\fi

\ifpaper
\begin{proposition}[$\ast$]
\else
\begin{proposition}
\fi
    \label{univ:prop:longestdescendingchain}
    Let $w \in \alphs$ and $\ell,n \in \N_0$. If $\alfrem{w^\ell} \supsetneq 
\ldots \supsetneq \alfrem{w^{\ell+n-1}}$ is a strictly descending chain of length 
$n$ then $n \leq \asize - 1$.
\end{proposition}
\ifpaper
\else
\input{proof_descendingchain}
\fi

\begin{remark}
The word $w = \ta_n \left[ \prod_{i=1}^\asize \ta_i^2 \right]^{n-2} \left[ 
    \prod_{i=1}^{\asize-1} \ta_i \right]$ over $\alphabet = 
\set{\ta_1,\dots,\ta_\asize}$ witnesses that a strictly descending chain of length 
$\asize - 1$ actually exists: we have  $\univ(w)=\asize-1$ and $\bdiff{\univ_w(s)} = 
\asize$ for all $s \in \disinv{1}{n-1}$ as well as $\bdiff{\univ_w(\asize)}= 
\asize-1$ and $\circuniv(w) = \asize-1$. Furthermore, $\alfrem{w} \supsetneq 
\ldots \supsetneq \alfrem{w^{\asize-1}}$ is a strictly descending chain of length 
$\asize-1$.
\end{remark}

With Proposition~\ref{univ:prop:longestdescendingchain} we can achieve our 
second main goal: the following theorem is a reasonable modification of 
Theorem~\ref{theo:barker22} such that its converse holds.

\begin{theorem}
    \label{theo:univ2}
    For $w \in \alphs$ the following statements are equivalent:\\
    1. $\bdiff{\univ_w(s)} = k+1$ for all $s \in \disinv{2}{\asize}$,\\
    2. $\bdiff{\univ_w(s)} = k+1$ for all $s \in \N_{\geq 2}$,\\
    3. $\circuniv(w) = k+1$.
\end{theorem}

\begin{proof}
    Firstly, let $\bdiff{\univ_w(s)} = k+1$ for all $s \in 
    \disinv{2}{\asize}$. Then Lemma~\ref{univ:lem:remsuf} gives us the 
    descending chain $\alfrem{w} \supseteq \ldots \supseteq \alfrem{w^\asize}$.
    By Lemma~\ref{univ:cor:longestascendingchain} this chain is not strict
    and thus, $\bdiff{\univ_w(s)} = k+1$ for all $s \in \N_{\geq 2}$. This proves 
the first implication. Now let $\bdiff{\univ_w(s)} = k+1$ for all $s \geq 2$.  
    Then, again, the chain $\alfrem{w} \supseteq \alfrem{w^2} 
    {\supseteq\ldots}$ is not strict. Hence the mapping $s \mapsto \rem(w^s)$ 
    is eventually constant. Thus there exists $t \geq 2$ such that $\rem(w^t) = 
    \rem(w^{t+1})$. Therefore, we have $\rem(\rem(w^t)w) = \rem(w^{t+1}) = 
    \rem(w^t)$. Note that, since $\bdiff{\univ_w(t+1)} = k+1$, we have 
    $\univ(\rem(w^t)w) = k+1$. Since, moreover, removing the remainder does not 
    change the universality of a word, $\univ(\rem(w^t) \cdot w \cdot 
    \rem(w^t)^{-1}) = k+1$ follows. Because the word $\rem(w^t) \cdot w \cdot 
    \rem(w^t)^{-1}$ is a conjugate of $w$, we have $\circuniv(w) = k+1$. This 
    proves the second implication. Finally, let $\circuniv(w) = k+1$. Then 
    Theorem~\ref{theo:barker22} implies immediately that $\bdiff{\univ_w(s)} = 
    k+1$ for all $s \in \disinv{2}{\asize}$.\qed
\end{proof}

Theorem~\ref{theo:univ2} provides an algorithm to compute the circular universality 
of a word $w$ in $\bigo(\sigma|w|)$, which is, if $\sigma<|w|$ holds, better than 
the na\"ive approach by computing $\iota(v)$ for every conjugate $v$ of $w$.

\ifpaper
\begin{proposition}[$\ast$]
\else
\begin{proposition}
\fi\label{runtime}
    Given a word $w \in \alphabet^{\ast}$, we can compute $\circuniv(w)$ 
in time $\bigo(\asize|w|)$.
\end{proposition}
\ifpaper
\else
\input{proof_runtime}
\fi

%% file: proof_remsuf.tex
\begin{proof}
Set $k=\univ(w)$.
    First note that $\rem(w^{s+1}) = \rem(\rem(w) w^s)$ holds. Second, let 
    $\ell = \univ(w^s)$ be the universality of $w^s$.  Now we examine the arch 
    factorisation of
    \[
    \rem(w) w^s = \rem(w) \cdot \prod_{i=1}^{\ell} \arch_i(w^s).
    \]
    For all $i \in \disinv{1}{\ell}$ there exist factors $x_i, y_i \in \alphs$ 
    such that the arches of $w^s$ are factorised by
    \begin{equation}
        \arch_i(w^s) = x_i y_i
    \end{equation}
    and such the arches of $\rem(w)w^s$ are factorised by
    \begin{equation}
        \arch_i(\rem(w) w^s) = \begin{cases}
            \rem(w) x_1, &\mbox{if $i = 1$,}\\
            y_{i-1} x_i, &\mbox{if $i \geq 2$}.
        \end{cases}
    \end{equation}
    Now the question is whether the remaining factor $y_{\ell} \rem(w^s)$ 
    contains yet another arch or whether it is already the remainder of 
    $\rem(w)w^s$.
    
    \textbf{Case 1:} Assume that it is already the remainder, i.e. that
    \[
    \rem(\rem(w)w^s) = y_{\ell} \rem(w^s)
    \]
    holds. Then $\rem(w)w^s$ has exactly $\ell$ arches and hence $w^{s+1}$ has 
    $k + \ell$ arches. In other words, $\bdiff{\univ_w(s+1)} = k$. But, since 
    $\rem(w^{s+1}) = \rem(w)w^s$, it also follows that $\rem(w^s)$ is a suffix 
    of $\rem(w^{s+1})$.
    
    \textbf{Case 2:} Now assume that $y_{\ell} \rem(w^s)$ contains yet another 
    arch. Then there exist factors $x_{\ell+1}, y_{\ell+1} \in \alphs$ such 
    that
    \begin{equation}\begin{aligned}
            \rem(w^s) &= x_{\ell+1} y_{\ell+1},\\
            \arch_{\ell+1}(\rem(w)w^s) &= y_{\ell} x_{\ell+1},\\
            \rem(\rem(w) w^s) &= y_{\ell+1}.
    \end{aligned}\end{equation}
    Then with similar arguments as in the previous case it follows that 
    $\bdiff{\univ_w(s+1)} = k+1$ and also that $\rem(w^{s+1}) = y_{\ell+1}$ is 
    a suffix of $\rem(w^s)$.
    
    And so we have either the case that $\bdiff{\univ_w(s+1)} = k$ and 
    $\rem(w^s)$ is a suffix of $\rem(w^{s+1})$, or the case that 
    $\bdiff{\univ_w(s+1)} = k+1$ and $\rem(w^{s+1})$ is a suffix of 
    $\rem(w^s)$.\qed
\end{proof}

%% file: proof_eventconstrict.tex
\begin{proof}
    From Corollary~\ref{univ:cor:chardiffrem} follows immediately that if $s 
    \mapsto \rem(w^s)$ is eventually constant then $\bdiff{\univ_w}$ is, too.
    
    So now assume that $\bdiff{\univ_w}$ is eventually constant. Then there is 
    an $s_0 \in \N$ such that $\bdiff{\univ_w(s)}=\ell$ for all $s \geq s_0$, 
    where $\ell$ is either $k$ or $k+1$.\\    
    \textbf{Case $\ell = k$:} Then Lemma~\ref{univ:lem:remsuf} implies that 
    there is an infinite ascending chain
    \[
    \alfrem{w^{s_0}} \subseteq \alfrem{w^{s_0+1}} \subseteq \ldots.
    \]
    But since $\alfrem{w^s} \subseteq \alphabet$ for all $s\in\N$ and 
    $\alphabet$ is finite, it follows that the chain is not strictly 
    increasing.  Hence there is some $t \geq s_0$ such that $\alfrem{w^t} = 
    \alfrem{w^{t+1}}$ and thus $s \mapsto \rem(w^s)$ is eventually constant.\\
    \textbf{Case $\ell = k+1$:} This case is analogous to the previous one.\qed
\end{proof}

%% file: proof_h.tex
\begin{proof}
    By the definition of the remainder we have $0 \leq \abs{\alfrem{u}} < 
    \asize$ for all $u \in \alphs$. Thus by
    \[
    0 \leq \abs{\alfrem{w^\ell}} < \ldots < \abs{\alfrem{w^{\ell+\asize-1}}} < 
    \asize
    \] 
    the claim follows.\qed
\end{proof}

%% file: proof_longestachain.tex
\begin{proof}
If there was a strictly ascending chain with length $\asize+1$ then 
Lemma~\ref{univ:lem:h} would imply $\asize-1 = \abs{\alfrem{w^{\ell+\asize-1}}} < 
\abs{\alfrem{w^{\ell+\asize}}}$ 
and thus $\abs{\alfrem{w^{\ell+\asize}}} = \asize$. This is a contradiction 
since not all letters can occur in the remainder. \qed
\end{proof}

%% file: proof_k.tex
\begin{proof}
    Let $\bdiff{\univ_w(s)} = k$ for all $s \in \disinv{1}{\asize}$. Then 
    Lemma~\ref{univ:lem:remsuf} gives us the ascending chain
    \[
    \alfrem{w^0} \subseteq \ldots \subseteq \alfrem{w^\asize}.
    \]
    By Corollary~\ref{univ:cor:longestascendingchain} this chain is not strict.  
    Therefore there exists an $s_0 \in \disinv{1}{\asize}$ such that 
    $\alfrem{w^{s_0-1}} = \alfrem{w^{s_0}}$. Then 
    Proposition~\ref{univ:prop:constant} implies that $s \mapsto \alfrem{w^s}$ 
    is eventually constant with $\alfrem{w^s} = \alfrem{w^{s_0-1}}$ for all $s 
    \geq s_0-1$. Consequently it follows from 
    Corollary~\ref{univ:cor:chardiffrem} that $\bdiff{\univ_w(s)} = 
    \bdiff{\univ_w(s_0)} = k$ for all $s \geq s_0$.\qed
\end{proof}

%% file: proof_1.tex
\begin{proof}
    Let $\ta\in\alphabet$ such that $\ta$ does not occur in $\rem(w)$. Then by 
    assumption $\ta$ occurs $k$ times in $w$ and therefore $sk$ times in 
    $w^s$.	So $\circuniv_w(s)$ is bounded by $sk$. However, it is also at 
    least $sk$.	Hence $\circuniv_w(s)=sk$.\qed
\end{proof}

%% file: proof_m.tex
\begin{proof}
    Because $\ta y \ta$ is a factor of $v$, there exist factors $x, 
    z\in\alphs$ such that $v=x\ta y \ta z$. Then
    \[
    u = \ta z x \ta y
    \]
    is a conjugate of $w$, too. Now suppose that $\bdiff{\univ_w}$ grew in the 
    $\asize^{\mathrm{th}}$ step by $k+1$, i.e.
    \begin{equation}
        \bdiff{\univ_w(\asize)} = k+1.
    \end{equation}
    Then the same holds for the circular universality of $u$ in at least some 
    step, i.e. $\bdiff{\circuniv_u(t)} = k+1$ for some $t \in \N$. This now 
    implies by Lemma~\ref{univ:lem:circtonorm} that the same must hold for its 
    plain universality, i.e. we have $\bdiff{\univ_u(t')} = k+1$ for some $t' 
    \in \N$. However, Proposition~\ref{univ:prop:k} states that any such growth 
    must have already occurred in the interval $\disinv{1}{\asize}$, i.e. there 
    is some $t'' \in \disinv{1}{\asize}$ with that property. But since by 
    assumption $\bdiff{\circuniv_w(s)} = k$ for all $s \in 
    \disinv{1}{\asize-1}$, the only possible value is $t'' = \asize$. Thus far 
    we have shown that \begin{equation}
        \label{eq:univ:lem:m:1}
        \bdiff{\univ_u(s)} = \begin{cases}
            k &\mbox{if $s \in \disinv{1}{\asize-1}$}\\
            k+1 & \mbox{if $s = \asize$}.
        \end{cases}
    \end{equation}
    The first case gives us by Lemma~\ref{univ:lem:remsuf} the ascending chain
    \[
    \alfrem{u^0} \subseteq \ldots \subseteq \alfrem{u^{\asize-1}}
    \]
    and the second case implies that this chain is strict. Then 
    Lemma~\ref{univ:lem:h} is applicable, resulting in $\abs{\alfrem{u^s}} = s$ 
    for all $s \in \disinv{1}{\asize-1}$. Now note that by 
    Lemma~\ref{lem:univ001} these arguments can be analogously applied to $u^R$ 
    as well. Consequently we can follow the same way that $\abs{\alfrem{(u^R)^s}} = 
    s$ for all $s \in \disinv{1}{\asize-1}$. And so we have in particular
    \begin{equation}
        \abs{\alfrem{u^{\asize-1}}} = \asize - 1
        \quad \mbox{and} \quad
        \abs{\alfrem{u^R}} = 1.
    \end{equation}
    Therefore, since by assumption $\abs{\alf(\ta y)} \leq \asize-1$, it 
    follows from the construction of $u$ that both
    \[
    \ta \in \alfrem{u^{\asize-1}}
    \quad \mbox{and} \quad
    \alfrem{u^R} = \set{\ta}.
    \]
    However, this implies that
    \[
    \alf(\rem(u^{\asize-1}) \rem(u^R)) = \alfrem{u^{\asize-1}} \neq \alphabet
    \]
    and hence we have $\bdiff{\univ_u(\asize)} = k$ by 
    Corollary~\ref{univ:cor:chardiffrem}. This is a contradiction to 
    Equation \eqref{eq:univ:lem:m:1} and so the supposition $\bdiff{\univ_w(\asize)} 
= 
    k+1$ must be false. Thus we have $\bdiff{\univ_w(s)}=k$ not only on the 
    interval $\disinv{1}{\asize-1}$, but on $\disinv{1}{\asize}$. Now the 
    claim follows by Proposition~\ref{univ:prop:k}.\qed
\end{proof}

%% file: proof_n.tex
\begin{proof}
    First let $\bdiff{\circuniv_w(s)} = k$ for all $s \in 
    \disinv{1}{\asize-1}$. If each letter occurs only once in each arch of $w$ 
    and at most once in the remainder $\rem(w)$, then the claim already 
    follows by Lemma~\ref{univ:lem:l}. So now assume that there exists some 
    letter $\ta \in \alphabet$ that occurs at least twice in an arch or the 
    remainder of $w$. Since the last letter of an arch is unique this implies 
    that the conditions of Lemma~\ref{univ:lem:m} are met. Then the claim 
    follows.
    
    The other direction follows immediately by Lemma~\ref{univ:lem:circtonorm}.\qed
\end{proof}

%% file: proof_i.tex
\begin{proof}
    The proof is symmetrical to the proof of Lemma~\ref{univ:lem:h}.\qed
\end{proof}

%% file: proof_equalsone.tex
\begin{proof}
    We can apply Lemma~\ref{univ:lem:i}. It implies that $\alfrem{w^\ell} = 
    \asize-1$, and hence $\ell \neq 0$. It also implies that $s \mapsto 
    \alfrem{w^s}$ assumes $\asize$ different values on the interval 
    $\disinv{\ell}{\ell+\asize-1}$, i.e. all possible values by 
    Lemma~\ref{univ:lem:boundalfrems}. Therefore $\alfrem{w^{\ell-1}}$ is one 
    of them, i.e. we have $\alfrem{w^{\ell-1}} = \alfrem{w^{\ell+m}}$ for some 
    $m \in \disinv{0}{\asize-1}$. However, by 
    Corollary~\ref{univ:cor:chardiffrem} their successors are also equal, i.e. 
    $\alfrem{w^{\ell}} = \alfrem{w^{\ell+m+1}}$. Hence, since the chain is 
    strict, $m = \asize-1$ is the only possible value for $m$. Thus, since 
    $\abs{\alfrem{w^{\ell+\asize-1}}}=0$ by Lemma~\ref{univ:lem:i}, it follows 
    that $\alfrem{w^{\ell-1}} = \emptyset$, and consequently
    \[
    \rem(w^{\ell-1}) = \emptyword = \rem(w^0).
    \]
    Then Lemma~\ref{univ:lem:eventperiodrem} implies that
    \[
        \rem(w^s) = \rem(w^{\ell-1+s})
    \]
    for all $s \in \N_0$, and therefore
    \[
    \alfrem{w} \supsetneq \ldots \supsetneq \alfrem{w^{\asize}}
    \]
    is a strictly descending chain of length $\asize$, too.\qed
\end{proof}

%% file: proof_descendingchain.tex
\begin{proof}
    We lead the proof by contradiction. It is structured as follows. First we 
    argue that one can assume $\ell = 1$. Second we argue that the conjugates 
    of $w$ obtained by cyclic shifts of whole arches give us strictly 
    descending chains, too. Third we follow that each arch and the remainder 
    end with the same letter. Last we show that this leads to a contradiction.
    
    Suppose that $n > \asize - 1$. Then in particular there is a strictly 
    descending chain of length $\asize$. By 
    Lemma~\ref{univ:lem:wlogellequalsone} we can assume w.l.o.g. that 
    $\ell = 1$. So we have
    \begin{equation}
        \alfrem{w} \supsetneq \ldots \supsetneq \alfrem{w^{\asize}}
    \end{equation}
    is a strictly descending chain.
    
    Now let $j \in \disinv{0}{k}$ and
    \begin{equation}
        w_j = \left[ \prod_{i=j+1}^k \arch_i(w) \right] \rem(w) \left[ 
        \prod_{i=1}^j \arch_i(w) \right].
    \end{equation}
    In other words, $w_j$ is a conjugate of $w$ obtained by cyclic shifts of 
    full arches of $w$ and in particular $w_0 = w$. Then for all $s\in\N$ the 
    arch factorisation
    \[
    w_j^s = \left[ \prod_{i=j+1}^{\univ(w^s)} \arch_i(w^s) \right] 
    \rem(w^s) \left[ \prod_{i=1}^j \arch_i(w^s) \right]
    \]
    contains at least $\univ(w^s)$ arches and thus
    \begin{equation}
        \univ(w_j^s) \geq \univ(w^s).
    \end{equation}
    Next, since $\alfrem{w^{\asize - 1}}$ is empty by Lemma~\ref{univ:lem:i}, 
    it follows that $\bdiff{\univ_w(\asize)} = k$, and so by 
    Theorem~\ref{theo:barker22} we must have $\circuniv(w) = k$. Hence by 
    Remark~\ref{univ:rem:boundtheuniversality}
    \[
    \univ(w_j^s) \leq \circuniv(w^s) \leq sk + s -1.
    \]
    for all $s \in \N$. However, because the chain is strict, 
    Corollary~\ref{univ:cor:sufremalphchain} implies that $\bdiff{\univ_w(s)} 
    = k+1$ on the interval $\disinv{2}{\asize}$ and thus $\univ(w^s) = sk + s 
    - 1$ for all $s\in\disinv{1}{\asize}$. Consequently we get for all $s \in 
    \disinv{1}{\asize}$ that
    \begin{equation}
        \univ(w_j^s) \leq \univ(w^s).
    \end{equation}
    But moreover note that, since $\rem(w^\asize) = \emptyword$, we have 
    $\univ(w_j^{\asize+1}) = \univ(w^{\asize+1})$ by definition of $w_j$. Thus 
    far we have shown that
    \begin{equation}
        \univ(w_j^s) = \univ(w^s)
    \end{equation}
    for all $s \in \disinv{1}{\asize+1}$. Then it follows for all $s \in 
    \disinv{2}{\asize}$ that $\bdiff{\univ(w_j^s)} = k+1$ and also that 
    $\bdiff{\univ(w_j^{\asize+1})} = k$. And so 
    Corollary~\ref{univ:cor:sufremalphchain} implies that
    \begin{equation}
        \alfrem{w_j} \supsetneq \ldots \supsetneq \alfrem{w_j^{\asize}}
    \end{equation}
    is also a strictly descending chain.
    
    Now we can apply Lemma~\ref{univ:lem:i} on this chain, too, and get
    \begin{equation}
        \abs{\alfrem{w_j}} = \asize - 1
        \quad\mbox{and}\quad
        \abs{\alfrem{w_j^{\asize-1}}} = 1.
    \end{equation}
    Since, moreover, $\bdiff{\univ(w_j^{\asize})} = k+1$ by 
    Corollary~\ref{univ:cor:chardiffrem} implies that
    \begin{equation}
        \alf(\rem(w_j^{\asize-1})\rem(w_j^R)) = \alphabet,
    \end{equation}
    it follows that
    \begin{equation}
        \abs{\alfrem{w_j^R}} = \asize - 1
        \quad\mbox{and}\quad
        \alfrem{w_j^{\asize-1}} \cap \alfrem{w_j^R} = \emptyset.
    \end{equation}
    Next let $\ta_j \in \alfrem{w_j^{\asize-1}}$ be the letter that occurs in 
    $\rem(w_j^{\asize-1})$. Note that $\ta_j$ is also the last letter of 
    $w_j$. Then, since $\alfrem{w_j^{\asize-1}}$ and $\alfrem{w_j^R}$ are 
    disjunct, we have
    \begin{equation}
        \label{eq:univ:lem:longestdescendingchain}
        \ta_j \notin \alfrem{w_j^R}.
    \end{equation}
    And so, since in $\rem(w_j^R)^R$ occur all letters of $\alphabet$ except 
    $\ta_j$ and it is also a prefix of the first arch of $w_j$, i.e. 
    $\arch_1(w_j)$, the last letter of $\arch_1(w_j)$ must be $\ta_j$, because 
    the last letter of an arch is unique. However, by construction of $w_j$ we 
    have that $\arch_1(w_j)$ is a suffix of $w_{j+1}$ for all $j < k$. This 
    implies that $\ta_j$ is the last letter of $w_{j+1}$ and hence $\ta_j = 
    \ta_{j+1}$ for all $j<k$. Therefore it follows inductively that
    \[
    \ta_j = \ta_0
    \]
    for all $j \leq k$. In other words, every arch and also the remainder of 
    $w$ ends with the same letter $\ta_0$.
    
    Now note that
    \[
    w_k = \rem(w) \left[ \prod_{i=1}^k \arch_i(w) \right].
    \]
    Then, since $\rem(w_k^R)^R$ and $\rem(w)$ are both prefixes of $w_k$ and 
    in both occur exactly $\asize-1$ different letters, it follows that
    \[
    \alfrem{w_k^R} = \alfrem{w}.
    \]
    But we have on the one hand $\ta_0 \in \alfrem{w}$, since $\ta_0$ is the 
    last letter of $w$, and on the other hand also $\ta_0 \notin 
    \alfrem{w_k^R}$ by \ref{eq:univ:lem:longestdescendingchain}. This is a 
    contradiction. Therefore the supposition $n > \asize - 1$ is wrong.\qed
\end{proof}

%% file: proof_runtime.tex
\begin{proof}
    By \cite[Proposition 10]{barker2020scattered} we can compute $\univ(w^\asize)$ 
in 
    $\bigo(\asize|w|)$. Let $k = \univ(w)$. If $\univ(w^\asize) = \asize k 
    + \asize - 1$, then $\circuniv(w) = k+1$, else $\circuniv(w) = k$.\qed
\end{proof}

%% file: conclusion.tex
The main goal of this work was to improve certain results from 
\cite{barker2020scattered} on the connection between the universality of 
repetitions and the circular universality, namely Theorem~\ref{theo:barker22} 
and Theorem~\ref{theo:barker23}. 

At first we focused our investigation on repetitions. In 
Section~\ref{univ:sec:powrem} we showed that the growth of the universality of 
repetitions can be characterised by their remainders and that the growth 
is eventually periodic beginning its periodicity latest after $\asize$ repetitions.
Thus, the universality of all other repetitions can be computed in constant time.
In Section~\ref{univ:sec:chaintherem} we found that one can translate questions 
about the universality of repetitions into questions about ascending or 
descending chains of the remainders of those 
repetitions. The investigation of 
strictly ascending chains led to a tight bound on the length of the 
longest possible strictly ascending chain and the connection of such chains with 
the circular universality, gives the extension of 
Theorem~\ref{theo:barker23} to alphabets of arbitrary size. On the other hand, on investigating
strictly descending chains, we found a tight bound on the length of such chains, which is 
surprisingly one step shorter than the ascending pendant. This lead to a modification of 
Theorem~\ref{theo:barker22} such that its converse holds, too, and also to an 
efficient algorithm to compute the circular universality of a word.

It remains an interesting open problem to 
characterise the class of words, for which the remainder of some proper 
repetition is the empty word. We propose to call such words \emph{perfect k-universal}. 
Furthermore, one could extend the study of $k$-universality from finite words 
to infinite words, e.g. one could study the universality of the sequence of 
finite prefixes of aperiodic infinite words.

%% file: appendix-proofs.tex
\noindent
\textbf{Proof of Lemma~\ref{univ:lem:circtonorm}.}

\input{proof_circtonorm}
\medskip

\noindent
\textbf{Proof of Lemma~\ref{univ:lem:prefix}.}

\input{proof_prefix}
\medskip

\noindent
\textbf{Proof of Proposition~\ref{univ:prop:chargrowth}.}

\input{proof_chargrowth}
\medskip

\noindent
\textbf{Proof of Corollary~\ref{univ:cor:chardiffrem}}

\input{proof_chardiffrem}
\medskip

\noindent
\textbf{Proof of Lemma~\ref{univ:lem:d}.}

\input{proof_d}
\medskip

\noindent
\textbf{Proof of Lemma~\ref{univ:lem:eventperiodrem}.}

\input{proof_eventperiodrem}
\medskip

\noindent
\textbf{Proof of Lemma~\ref{univ:prop:eventperiodalfrem}.}

\input{proof_eventperiodalfrem}
\medskip

\noindent
\textbf{Proof of Proposition~\ref{evperiodic}.}

\input{proof_evperiodic}
\medskip

\noindent
\textbf{Proof of Lemma~\ref{univ:lem:boundalfrems}.}

\input{proof_boundalfrems}
\medskip

\noindent
\textbf{Proof of Theorem~\ref{univ:theo:periodicity}.}

\input{proof_periodicity}
\medskip

\noindent
\textbf{Proof of Proposition~\ref{division}.}

\input{proof_division}
\medskip

\noindent
\textbf{Proof of Propoisition~\ref{consttime}.}

\input{proof_consttime}
\medskip

\noindent
\textbf{Proof of Lemma~\ref{univ:lem:remsuf}.}

\input{proof_remsuf}
\newpage

\noindent
\textbf{Proof of Lemma~\ref{univ:lem:eventconstcrit}.}

\input{proof_eventconstrict}
\medskip

\noindent
\textbf{Proof of Lemma~\ref{univ:lem:h}.}

\input{proof_h}
\medskip

\noindent
\textbf{Proof of Corollary~\ref{univ:cor:longestascendingchain}.}

\input{proof_longestachain}
\medskip

\noindent
\textbf{Proof of Proposition~\ref{univ:prop:k}.}

\input{proof_k}
\medskip

\noindent
\textbf{Proof of Lemma~\ref{univ:lem:l}.}

\input{proof_1}
\newpage

\noindent
\textbf{Proof of Lemma~\ref{univ:lem:m}.}

\input{proof_m}
\newpage

\noindent
\textbf{Proof of Proposition~\ref{univ:theo:n}.}

\input{proof_n}
\medskip

\noindent
\textbf{Proof of Lemma~\ref{univ:lem:i}.}

\input{proof_i}
\medskip

\noindent
\textbf{Proof of Lemma~\ref{univ:lem:wlogellequalsone}.}

\input{proof_equalsone}
\medskip

\noindent
\textbf{Proof of Proposition~\ref{univ:prop:longestdescendingchain}.}

\input{proof_descendingchain}
\medskip

\noindent
\textbf{Proof of Proposition~\ref{runtime}.}

\input{proof_runtime}